\documentclass{article} 
\usepackage{arxiv}

\usepackage{arydshln}
\usepackage{makecell}

\usepackage{algorithmic}
\usepackage{graphicx}
\usepackage{textcomp}
\usepackage{xcolor}

\usepackage{latexsym}
\usepackage{verbatim}

\usepackage{times}
\usepackage{epsfig}
\usepackage{array, multirow}
\usepackage{nicematrix}
\usepackage{pifont}

\usepackage{amssymb}
\usepackage{amsmath}
\usepackage{amsthm}
\usepackage{amsfonts}
\usepackage{booktabs}
\usepackage{enumitem}
\usepackage{color}

\usepackage[utf8]{inputenc} 
\usepackage[T1]{fontenc}    
\usepackage{hyperref}       
\usepackage{url}            
\usepackage{amsfonts}       
\usepackage{nicefrac}       
\usepackage{microtype}      
\usepackage{cleveref}       
\usepackage{natbib}
\usepackage{doi}
\usepackage{authblk}

\def\BibTeX{{\rm B\kern-.05em{\sc i\kern-.025em b}\kern-.08em
    T\kern-.1667em\lower.7ex\hbox{E}\kern-.125emX}}

\newcommand{\cmark}{\textcolor{green}{\ding{51}}}
\newcommand{\xmark}{\textcolor{red}{\ding{55}}}
\newcommand{\frozen}{\raisebox{-0.25em}{\includegraphics[height=1em]{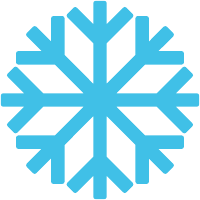}}}
\newcommand{\trainable}{\raisebox{-0.25em}{\includegraphics[height=1em]{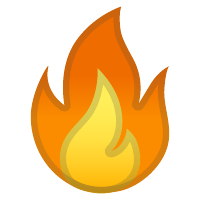}}}
\newcommand{\blfootnote[1]}{\begingroup\renewcommand\thefootnote{}\footnote{#1}\addtocounter{footnote}{-1}\endgroup}

\makeatletter
\def\@fnsymbol#1{\ensuremath{\ifcase#1\or \dagger\or \ddagger\or
   \mathsection\or \mathparagraph\or \|\or **\or \dagger\dagger
   \or \ddagger\ddagger \else\@ctrerr\fi}}
\def\blfootnote{\gdef\@thefnmark{}\@footnotetext}
\def\astfootnote{\gdef\@thefnmark{$*$}\@footnotetext}
\makeatother

\title{MuJo: Multimodal Joint Feature Space Learning for Human Activity Recognition}

\author[1,*]{Stefan Gerd Fritsch}
\author[2,*]{Cennet Oguz}
\author[1,3,*]{Vitor Fortes Rey}
\author[1]{Lala Ray}
\author[1,3,4]{Maximilian Kiefer-Emmanouilidis}
\author[1,3]{Paul Lukowicz}

\affil[1]{Embedded Intelligence, DFKI, 67663 Kaiserslautern, Germany}
\affil[2]{Multilinguality and Language Technology, DFKI, 66123 Saarbr\"ucken, Germany}
\affil[3]{Department of Computer Science, RPTU Kaiserslautern-Landau, 67663 Kaiserslautern, Germany}
\affil[4]{Department of Physics, RPTU Kaiserslautern-Landau, 67663 Kaiserslautern, Germany}
\affil[ ]{\textbf{Email:} \texttt{firstname.lastname@dfki.de}}
\affil[*]{Equal contribution.}
\date{}

\renewcommand{\headeright}{}
\renewcommand{\undertitle}{}

\hypersetup{
pdftitle={MuJo: Multimodal Joint Feature Space Learning for Human Activity Recognition},
pdfsubject={cs.LG},
pdfauthor={Stefan Gerd~Fritsch, Cennet~Oguz, Vitor~Fortes Rey, Lala~Ray, Maximilian~Kiefer-Emmanouilidis, Paul~Lukowicz},
pdfkeywords={multimodal representation learning, human activity recognition, multimodal fitness dataset},
}
\begin{document}
\maketitle

\begin{abstract}
Human activity recognition (HAR) is a long-standing problem in artificial intelligence with applications in a broad range of areas, including healthcare, sports and fitness, security, and more. The performance of HAR in real-world settings is strongly dependent on the type and quality of the input signal that can be acquired. Given an unobstructed, high-quality camera view of a scene, computer vision systems, in particular in conjunction with foundation models, can today fairly reliably distinguish complex activities. On the other hand, recognition using modalities such as wearable sensors (which are often more broadly available, e.g., in mobile phones and smartwatches) is a more difficult problem, as the signals often contain less information and labeled training data is more difficult to acquire.
To alleviate the need for labeled data, we introduce our comprehensive Fitness Multimodal Activity Dataset (FiMAD) in this work, which can be used with the proposed pre-training method MuJo (Multimodal Joint Feature Space Learning) to enhance HAR performance across various modalities. FiMAD was created using YouTube fitness videos and contains parallel video, language, pose, and simulated IMU sensor data. MuJo utilizes this dataset to learn a joint feature space for these modalities. We show that classifiers pre-trained on FiMAD can increase the performance on real HAR datasets such as MM-Fit, MyoGym, MotionSense, and MHEALTH. For instance, on MM-Fit, we achieve a Macro \( F_1 \)-Score of up to 0.855 when fine-tuning on only 2\% of the training data and 0.942 when utilizing the complete training set for classification tasks. We compare our approach with other self-supervised ones and show that, unlike them, ours consistently improves compared to the baseline network performance while also providing better data efficiency.
\end{abstract}

\begin{NoHyper}
\blfootnote{This research was supported by the XAINES project, funded by the German Federal Ministry of Education and Research under grant agreement No 01IW20005, and by the SustainML project, funded by the European Union’s Horizon Europe research and innovation programme under grant agreement No 101070408.}
\setcounter{footnote}{0}
\end{NoHyper}

\section{Introduction}
\begin{figure*}[!t]
\centering
\includegraphics[width=0.9\textwidth]{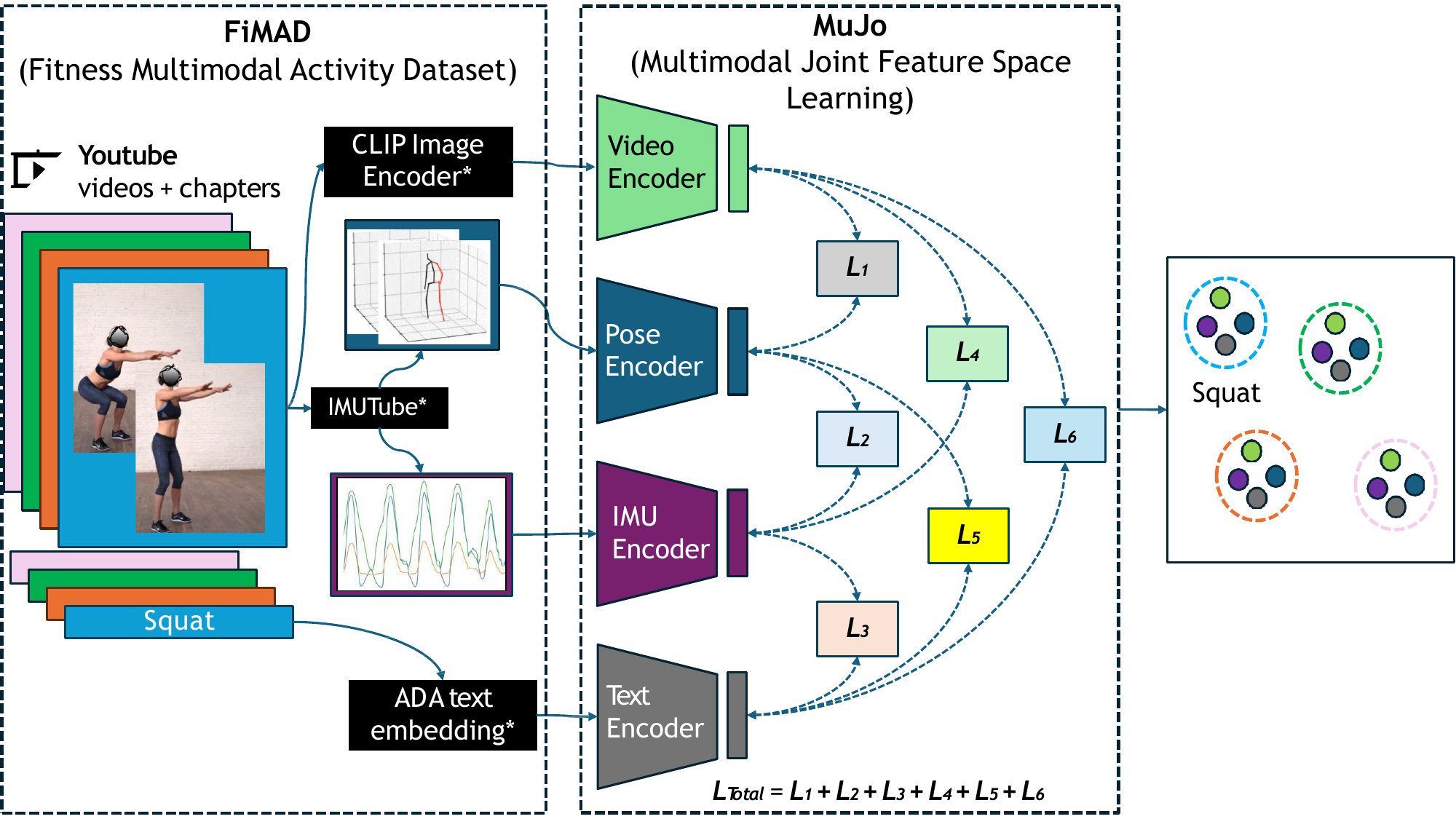}
\caption{\centering The pipeline depicts the construction process of FiMAD and the training of MuJo for multimodal joint feature space learning. The asterisk (*) indicates that the input is being pre-calculated (frozen) and not optimized during the training process.}
\label{fig:1}
\end{figure*}

 Perceiving and interpreting human activities is a core functionality of intelligent systems with applications in a wide range of areas, including healthcare~\citep{tan2021human}, sports and fitness~\citep{host2022overview,nadeem2020accurate}, and security~\citep{sunil2021usual}, among others. A basic capability needed to facilitate such functionality is the identification of concrete human actions from the system's sensory input, referred to as human activity recognition (HAR). In general, HAR is a difficult problem, especially when applied to uncontrolled real-life environments, where it must contend with the variability of human actions and the open-ended variations in environmental conditions. In such settings, the performance of HAR systems is strongly dependent on the available input modalities. Recent advances in computer vision, particularly with large vision-language models (LVLM)~\citep{radford2021learning,achiam2023gpt}, have enabled impressive performance when provided with sufficiently high-quality images. In many cases, simply feeding images to publicly available tools such as GPT-4 can result in accurate descriptions of scenes that involve complex and subtle activities. However, in real-life settings, such images are not always available; either due to a lack of cameras or because of privacy concerns. Much more broadly available input modalities are wearable sensors, in particular, inertial measurement units (IMUs), which are ubiquitous in smartphones, smartwatches, or even earbuds, all of which people carry with them on a daily basis. In contrast to cameras, such sensors are more privacy-preserving, since they collect only data from the subject who wears them. Moreover, they are cost- and energy-efficient and, for instance, in the case of smartwatches and smartphones, convenient to wear. Unfortunately, achieving acceptable recognition performance with such sensors is a much more challenging problem due to two factors, the first being \textbf{information ambiguity}. An IMU (a motion sensor) in a smartwatch on the wrist or in a smartphone in a pocket contains much less detailed information about the user's activity than a high-quality picture. This problem is heightened by the second factor, \textbf{the lack of training data}.
 IMU-based HAR datasets are very small compared to vision HAR ones. For example, the total length of all sensor-based HAR datasets in the popular UCI machine learning repository~\citep{kelly2025uci} is less than 63 hours~\citep{rey2021translating}, which pales in comparison to a single video HAR dataset such as Kinetics-700 with around 1800 hours~\citep{carreira2019short}. Moreover, the massive amount of text/image and text/video data pairs available online led to the success of LVLMs, which can be used for training aligned representations that capture the semantic similarity of multiple modalities. 
 On the other hand, although large amounts of unlabeled IMU data are increasingly becoming available~\citep{chan2024capture24}, labeled training data remain scarce.
 The work presented in this paper, which aims to address these issues, is based on three key observations:
 \begin{enumerate}
     \item Although the availability of input modalities during inference, the time when the system is utilized in the real world, may be constrained, the input modalities during training can be selected with much greater flexibility.
     \item Representation learning facilitates knowledge transfer between modalities by aligning embedding spaces, in particular when synchronized training data is available. 
     \item Significant progress has recently been made in generating synthetic data for a variety of sensors from videos. For example, IMUTube~\citep{kwon2021imutube2} generates virtual gyroscope and accelerometer data from videos, while pose extraction methods such as OpenPose~\citep{cao2017realtime} and VideoPose3D~\citep{pavllo20193d} are used to generate skeletal data of human activities. This enables the creation of synchronized video/sensor training data, and through the alignment of video with text, it facilitates the synchronization of sensor data with textual descriptions.
 \end{enumerate}
 Based on the above considerations, we propose a Multimodal Joint Feature Space Learning method (MuJo) that learns a joint representation between videos, video-derived poses, virtual IMU data generated from videos, and textual descriptions by employing pair-wise contrastive learning, as depicted in Figure \ref{fig:1}. MuJo aims to utilize the complementary information of each modality to improve the performance for both unimodal and multimodal inference. In particular, we show that our method dramatically reduces the need for labeled training data for downstream tasks. For that, we provide an extensive evaluation on four publicly available HAR datasets, namely MM-Fit~\citep{stromback2020mmfit}, MyoGym~\citep{koskimaki2017myogym}, MotionSense~\citep{malekzadeh2018motionsense}, and MHEALTH~\citep{banos2015mhealth}. For instance, on the MM-Fit dataset in the multimodal case (encompassing all the aforementioned modalities as input), our method improves the recognition rate (Macro $F_1$-Score) from 0.897 to 0.942 when using all available training data. Even when utilizing only 2\% of the training data, it enhances the recognition rate from 0.790 to 0.855.
 
Overall, the key contributions of this work can be summarized as follows:

\begin{enumerate}
    \item Development of the Fitness Multimodal Activity Dataset (FiMAD)\footnote{\url{https://github.com/R-tree/MuJo.git}\label{github}}, a comprehensive dataset containing parallel video, language (including video clip labels, instructor utterances, and detailed exercise descriptions generated by GPT-3.5 Turbo), pose data, and simulated IMU data points.
    \item Pre-training of MuJo\footref{github}, a multimodal joint space model, using the FiMAD dataset. MuJo integrates four different modalities: text, video, pose, and virtual sensor data, leveraging contrastive learning inspired by CLIP~\citep{radford2021learning}.
    \item Evaluation of the utility and generalizability of MuJo for real-world accelerometer data on the MM-Fit, MyoGym, MotionSense, and MHEALTH datasets. On the MM-Fit dataset, we additionally conducted experiments with pose, video, and multimodal inputs. The benefits of pre-training MuJo on FiMAD, especially when limited data are available for the target task, are analyzed. Furthermore, we analyze the zero-shot performance of our method and compare unimodal versus multimodal training on all four datasets.
\end{enumerate}

\section{Related Work}

\paragraph{Datasets for activity recognition}

In recent years, several datasets have been developed to advance HAR for sports activities, supporting various applications and enabling robust evaluations across multiple modalities, including images~\citep{verma2020yoga}, videos~\citep{shao2020finegym}, and wearable sensors~\citep{malekzadeh2018motionsense}. Furthermore, cross-modal datasets, such as video-pose~\citep{fieraru2021aifit} and video-sensor datasets~\citep{stromback2020mmfit,kwon2021complex}, have been developed to support transfer learning in HAR~\citep{thukral2023cross,kamboj2024survey}, allowing knowledge transfer between different data modalities and enhancing model adaptability. However, these datasets are dependent on human-annotated activity labels, a process that is both costly and time-consuming, posing challenges for large-scale data collection and scalability. For example, in~\citep{kwon2021complex}, virtual IMU data was simulated from YouTube videos and combined with real data to improve classification performance. However, this approach required collecting and labeling videos for target activities as well as careful calibration of the generated simulated sensor data. Therefore, we propose a novel data curation method that utilizes YouTube chapters to collect videos annotated with corresponding activity classes, while also expanding the range of modalities (explained in Section \ref{dataset_construction}). This method facilitates the collection of multimodal data encompassing sensor, pose, video, and text features to enable learning of a richer vector space.

\paragraph{Unimodal and multimodal activity recognition}
Unimodal activity recognition has been extensively studied in HAR. Various approaches have been proposed to address this task and can be broadly categorized into two groups: handcrafted feature-based and deep learning-based methods.
Handcrafted feature-based methods rely on extracting low-level features from the input data and designing a classifier to recognize activities. For instance, in accelerometer-based activity recognition~\citep{kwapisz2011activity}, features such as the mean, standard deviation, and energy of the acceleration signal are often used to capture the characteristics of different activities. Popular classifiers used in this approach include decision trees, support vector machines, and k-nearest neighbors.
Deep learning-based methods, on the other hand, use neural networks to learn features and recognize activities from raw input data. Convolutional neural networks (CNNs) and recurrent neural networks (RNNs) are two popular types of neural networks used in this approach. CNNs are commonly used for image-based activity recognition~\citep{stromback2020mmfit}, being combined with RNNs in some cases~\citep{ordonez2016deep}.

Despite the success of unimodal activity recognition, it is often limited because a single modality may only capture a part of the essential information about the activity, leading to lower recognition accuracy. Therefore, multimodal activity recognition, which combines information from multiple sources, such as IMU data, audio, and video~\citep{mekruksavanich2022multimodal,ijaz2022multimodal}, has gained increasing attention in recent years.
These modalities can be combined using fusion methods such as early or late fusion~\citep{duhme2022fusion}.
Some studies have also focused on domain adaptation, where the goal is to recognize activities in a new environment using data from a different environment~\citep{plananamente2022test}. 
In such cases, transfer learning techniques can adapt the model to the new environment using a small amount of labeled data.
Overall, multimodal activity recognition has shown promising results, with improved accuracy compared to unimodal approaches. However, challenges in data fusion, feature extraction, and model selection still need to be addressed to improve the performance of multimodal activity recognition systems~\citep{islam2022human}.

\paragraph{Representation learning with proxy tasks} Different modalities, e.g., natural language, visual inputs, or sensor signals, often complement a common concept. A key challenge of multimodal feature representation is exploring efficient methods for multimodal fusion with the given modalities to avoid missing modal features. We can differentiate the representation learning methods in two aspects: learning objectives and fusing the learned features in a joint space. In learning objectives, various approaches have been proposed to learn multimodal representations by formulating proxy tasks such as reconstruction, completion~\citep{seo2021look,sun2019videobert}, matching/alignment~\citep{li2020hero,miech2020end}, and ordering tasks~\citep{li2020hero}. Generating one feature vector with multimodal inputs for a common concept is a crucial problem of multimodal representation learning. Simple methods, such as concatenating learned unimodal feature vectors, have been extensively explored to enhance the performance of downstream tasks~\citep{hu2018early,stromback2020mmfit}. These methods involve combining feature vectors from a single modality, such as text or image, to create more informative representations that can be used for various tasks. 
More advanced methods have also been proposed, including a tensor fusion network~\citep{zadeh2017tensor} and a transformer-based co-attention mechanism for efficient cross-modality learning~\citep{lu2019vilbert}.  

Proxy tasks were also developed for the field of HAR with wearable devices including IMUs, where unlabeled IMU data are used to pre-train better representations. 
Those approaches include Multitask~\citep{multitask}, Auto-encoder~\citep{reconstruction} and simCLR~\citep{simclr}.  Multi-task~\citep{multitask} proposes auxiliary tasks of binary classification for extracting useful features for the downstream task. A multitask temporal CNN is trained to classify whether the input signal was transformed or not for each of a set of transformations such as scaling, rotating the signal, etc. The authors utilize a large unlabeled accelerometer dataset for self-supervised learning and transfer the learned knowledge to a small, labeled activity recognition dataset. Auto-encoder~\citep{reconstruction} is the application of auto-encoders, where an encoder network reduces the dimensionality of the data, while a decoder brings the representation back to its original dimensions, with the whole system learning to reconstruct the original signal. simCLR~\citep{simclr} learns general representations from a dataset by training a model to match different view instances created from the same signal by contrasting them. The positive pair is obtained by the transformation of the input window, while the negative pair is formed by the remaining input windows of the same batch. Then, the contrastive loss pulls the positive pair together, while pulling the negative pair far away from each other in the representation space.

\paragraph{Representation via contrastive learning}
Contrastive learning is an efficient self-supervised framework applied across multiple domains, which learns similar/dissimilar representations from data that are organized into similar and dissimilar pairs. Recently, Radford et al. presented a novel approach, known as CLIP~\citep{radford2021learning}, to jointly represent images and text descriptions by leveraging contrastive loss with the similarity of image-text pairs. CLIP learns two feature spaces with two encoders, i.e., image and text encoders, then projects the learned features into a shared latent space with contrastive loss.
Contrastive learning has also been extended to other modalities. For example, video-text data have attracted interest in studies focused on the shared space of video-text features~\citep{xu2021videoclip,xue2022clip} and text-to-video generation~\citep{singer2022make}. MotionCLIP~\citep{tevet2022motionclip} trained an encoder to find the appropriate embedding of an input sequence in the CLIP space and a decoder to generate the most fitting motion for a given CLIP space. Following this, IMU2CLIP~\citep{moon2022imu2clip} was proposed to learn a joint space that aligns IMU motion sensor recordings with video and text by projecting them into a shared representation space. 
Studies in action recognition~\citep{li2020hero,wang2021actionclip,sun2022human,morshed2023human} have also significantly enhanced the field of fitness activity recognition, despite traditionally focusing on a broad spectrum of actions such as ``cutting the potato'' and ``cleaning the drawer''. Certain investigations in action recognition also aim to acquire knowledge of the feature space through contrastive learning, as exemplified by initiatives such as HERO~\citep{li2020hero} and ActionCLIP~\citep{wang2021actionclip}.

The achievements discussed in the aforementioned studies have shown that contrastive learning has a significant impact on extracting a joint space with similar and dissimilar pairs. 
Consequently, the contrastive learning methodology employed in our study aims to acquire knowledge of the feature space pre-training to fitness activities. Our multimodal contrastive learning includes video, IMU sensors, poses, and text, within a multitask training manner in an end-to-end training paradigm similar to ImageBind~\citep{girdhar2023imagebind}. 
Recent works in multimodal pre-training also make use of simulated sensor data from video sources~\citep{fortes2024enhancing}, but focus only on sensors placed on the wrist and do not include video features. Moreover, they do not investigate pre-training on a large amount of YouTube fitness data, relying instead on sign language videos and motion capture datasets.

\section{Dataset Collection and Construction}\label{dataset_construction}
Annotating video descriptions is time-consuming and expensive. Thus, we created our dataset from a comprehensive and varied collection of videos with temporally localized chapter information, including chapter titles and start times, specifically for fitness activities. YouTube provides chapters within a video, each with its preview, providing information and context for different parts of the video. Those chapters are manually added by the uploader or automatically generated.
Each chapter is a continuous, non-overlapping segment that divides the video into a video clip and its corresponding textual description.

We manually collected 1388 fitness videos with pre-annotated video chapters. Additionally, we retrieved the captions automatically generated by YouTube. All our videos show a tutor who teaches how to do fitness exercises (e.g., ``squat with dumbbells'', ``sit-ups'', ``side deep squats'', etc.). The trainers primarily conduct HIIT home workouts, focusing on bodyweight exercises with little to no equipment. The exercise videos provide instructions for the activities while demonstrating them clearly. The annotation does not include predefined categories or labels as each video owner annotated the activity segment of the videos with their personal preferences, such as ``deep squats with dumbbells'', ``high jump squats'', etc., instead of only ``squats''. Based on the temporal boundaries of the chapters within each video, we split the videos into shorter activity clips along with their corresponding caption segments. Since workout exercises are typically short, activity videos that are too long often contain multiple exercises and, thus, incorrect chapter annotations. Hence, we only kept exercises that are shorter than 2 minutes and discarded longer exercises. After that, the dataset consists of 10,695 (video, chapter, caption)-samples.

We converted all 10,695 exercise videos to 720p resolution and 50 frames per second via FFmpeg's adaptive overlapped block motion compensation with bidirectional motion estimation. To enhance the dataset with diverse modalities, we used IMUTube~\citep{kwon2021imutube2} to obtain poses and virtual sensor (accelerometer and gyroscope) features. As a result, our dataset includes (video, pose, sensor, text)-pairs for each input. For video and text, we extracted multiple features. The video features were generated using HERO~\citep{li2020hero}, using CLIP~\citep{radford2021learning} and S3D~\citep{xie2018rethinking} embeddings. For the text modality, we applied several normalization strategies to the raw chapters, such as replacing abbreviations within the text (e.g., changing ``l'' to ``left'' and ``r'' to ``right'') as well as removing stop words, numbers, and some special symbols from a predefined list. In addition, we used GPT-3.5 Turbo to generate detailed descriptions of the exercises. We then extracted multiple text features with various models, including OpenAI's text-embedding-ada-002 embedding model (ADA), CLIP, BERT, as well as language features and motion tokens extracted with the Text2Motion Transformer~\citep{guo2022generating}. We applied IMUTube to generate virtual gyroscope and accelerometer data for sensors placed on all 16 joint nodes. We also used the implementation of IMUTube to extract 3D poses from the videos, which utilizes OpenPose followed by VideoPose3D for 3D pose lifting. Since IMUTube failed to extract sensor data for 884 exercise videos, we excluded these sequences. We used the remaining 9,811 exercises as our dataset and split it into 8,227 train and 1,584 validation activity clips. The split is based on the level of the 1,388 original fitness videos to ensure video independence. We did not create a test set because the dataset is used solely for pre-training, not for testing. Since workout session videos typically start and end with explanations given by the trainer, we discarded the initial and final 15\%\footnote{This value has proven to be effective in experimental investigations.} of each activity clip. Despite these filtering steps, we receive a large dataset totaling around 78 hours in length. Instances were generated from the videos using sliding windows $x_i$ of 100 frames in length (i.e., two seconds) with an overlap of 50 frames. We found this size suitable for fitness exercises, as a two-second window generally captures enough meaningful information regarding the subject's movements.

In summary, our dataset comprises a large collection of short fitness activity clips containing video features, poses, virtual IMU data, chapter labels, captions, and GPT-generated descriptions, as well as several types of features extracted from these text modalities.

\section{Method}
Our method leverages multimodal information for HAR pre-training. In this section, we outline the task and its formulation, along with detailing the multimodal representation methods employing CLIP~\citep{radford2021learning}-like contrastive learning. Our objective is to improve HAR by employing contrastive learning across multiple modalities. The fundamental idea is that the information extracted from different modalities for a short video interval should be similar, as they represent the same activity from various perspectives.

Each instance is represented as a window, and for each window, we extract features for every modality. This means that our data consists of $\mathcal{X}^m = \{x^1_m, ..., x^N_m\}$, where $x^i_m$ is the feature of modality $m$ in window $i$. The modalities we utilize include:
\begin{itemize}
    \item $\mathcal{X}^{pose}$: The sequence of 3D poses for the subject extracted with ImuTube~\citep{kwon2021imutube2} utilizing OpenPose~\citep{cao2017realtime} and VideoPose3D~\citep{pavllo20193d}.
    \item $\mathcal{X}^{video}$: Video features extracted with HERO~\citep{li2020hero} using CLIP~\citep{radford2021learning} embeddings.
    \item $\mathcal{X}^{acc}$: Simulated accelerometer data generated by ImuTube.
    \item $\mathcal{X}^{text}$: Embeddings extracted using OpenAI's ADA model from YouTube's chapter information with a lightweight normalization process removing a few special symbols and replacing abbreviations.
\end{itemize}
Each modality $m$ has its own encoder network $e_m$ and projection layers $p_m$. Thus, $rep_m(x_m) = p_m(e_m(v_m))$ serves as our representation for a vector $v_m$ of modality $m$ in our shared representation space.
We learn this shared space between the modality representations using the InfoNCE loss. First, for all possible  sets of two modalities $\{m_a, m_b\}$, we compute similarities using 
\begin{equation}
    {sim(v_a, v_b)} = \frac{{rep_{m_a}(v_a)} \times {rep_{m_b}(v_b)}^\top}{\lVert {rep_{m_a}(v_a)} \rVert \cdot \lVert {rep_{m_b}(v_b)} \rVert}
\end{equation}
Then, those similarities are softmax-normalized within the batch to calculate the loss between the two modalities.
\begin{equation}
\mathcal{L}_\text{c}(x_a,x_b) = -\sum_{i} \log\frac{\text{exp}(sim(x^i_a,x^i_b)/\tau)}{\sum_{j}\text{exp}(sim(x^i_a,x^j_b)/\tau)},
\end{equation}
where $x_a$ and $x_b$ correspond to aligned batch data for modalities $a$ and $b$, respectively, and $\tau$ is the temperature hyper-parameter \footnote{Best results obtained with $\tau=1.0$.}. In other words, for each data instance, its representations across different modalities (positive) should be more similar to each other than at other instances (negatives, which are all other examples in the same batch). As in CLIP, we compute $\mathcal{L}_c(x_{m_a},x_{m_b}) + \mathcal{L}_c(x_{m_b},x_{m_a})$ for each pair, symmetrically aligning the embeddings and covering the negatives from both perspectives. Since we calculate losses between all modalities,
our total loss amounts to
\begin{equation}
   \mathcal{L}_{total} = \frac{1}{\binom{M}{2}} \sum_{m_a,m_b} (\mathcal{L}_c(x_{m_a},x_{m_b}) + \mathcal{L}_c(x_{m_b},x_{m_a}) ),
\end{equation}
where $M$ is the total number of modalities. In our case, four modalities generate six combinations, as shown in Figure \ref{fig:1}.

\paragraph*{Hyperparameter search and model optimization}
We applied a random search approach to identify the optimal hyperparameters for our model. Our hyperparameters included the number and size of encoder and projection layers, learning rate, dropout rates, and activation functions. Furthermore, we experimented with various configurations of the loss function, specifically comparing the effect of using the text modality as an anchor, i.e., only considering sensor-text, pose-text, and video-text loss pairs, versus incorporating all possible modality combinations into the loss. 
Encompassing all modality pairs together yielded superior results. We explored weighting the text modality higher than the others in the loss, but it did not have any observable effect on performance. We also explored the use of S3D-based video embeddings, the gyroscope as the sensor input, and the other text modalities and features described in Section \ref{dataset_construction} as inputs. However, since those configurations did not yield optimal performance during the hyperparameter search, we have omitted those results for the sake of brevity.

\paragraph*{Model structure}
Based on the results of the hyperparameter tuning, we designed our encoder structures to include a model with a single fully connected layer for $e_{text}$ and two fully connected layers for $e_{video}$. $e_{text}$ has an input size of 1536 (the size of the ADA text embeddings) and an output size of 768. $e_{video}$ has an input size of 1024 (size of the HERO CLIP features), a hidden layer size of 256, and an output size of 256. For $e_{pose}$ and $e_{sensor}$, we employed the CNN encoder architecture as described in~\citep{stromback2020mmfit}, making slight adaptations to accommodate our input shapes. Both encoders consist of three CNN layers. $e_{pose}$ has a kernel size of (11, 11) and includes two fully connected layers, each with a size of 1024. $e_{sensor}$ has a kernel size of ($s_n$, 11) for the first layer, where $s_n$ represents the number of sensor inputs. The kernel sizes for layer 2 and layer 3 are (1, 11), followed by a single fully connected layer with an output size of 512.
The projections $p_m$ are fully connected three-layer models for all modalities $m$, with hidden and output layer sizes of 1280. All encoders and projections utilize GELU as activation function and dropouts of 0.4 for the projections, 0.3 for $e_{video}$, 0.6 for $e_{text}$, 0.4 for $e_{pose}$, and no dropout for $e_{sensor}$.
We trained with a batch size of 256 for a maximum of 200 epochs with early stopping (patience = 50). We utilized the AdamW optimizer with a learning rate of 0.0008 in combination with a cosine annealing scheduling for the learning rate during training that starts with 34 warm up epochs followed by a maximum number of 33 iterations using a minimum learning rate of 3.0398e-06.

\section{Evaluation Methodology}
\subsection{Evaluation Datasets}
We conducted experiments on the MM-Fit~\citep{stromback2020mmfit}, MyoGym~\citep{koskimaki2017myogym}, MotionSense~\citep{malekzadeh2018motionsense}, and MHEALTH~\citep{banos2015mhealth} datasets. MM-Fit is designed for multimodal recognition of fitness activities and combines data from multiple modalities, including video, pose estimations, and various sensors. The dataset includes 21 workout sessions conducted by ten subjects, each performing ten different fitness exercises.
The MyoGym dataset comprises 30 distinct gym exercises focused on upper body training by ten individuals, each equipped with a sensor on their right forearm.
MotionSense consists of sensor data collected from 24 participants carrying an iPhone 6s in their front pocket, covering six different activities (walking downstairs, walking upstairs, sitting, standing, and jogging).
The MHEALTH dataset includes twelve physical activities (such as climbing stairs, cycling, and running) performed by ten subjects, with data captured by three sensors positioned on the chest, right wrist, and left ankle.
All datasets, except for MotionSense, include a ``No Activity'' category (NULL class) in addition to the specified activities.

\subsection{Unimodal Classification on Real-World Accelerometer Data}
\label{um-class}
We evaluated the performance of our pre-trained representation on real-world accelerometer data using all four datasets. For MM-Fit, we used the accelerometer data from the smartwatches worn on the left and right wrists, which we had to downsample from 100 Hz to 50 Hz via linear interpolation to match our input frequency. For MyoGym, MotionSense, and MHEALTH, we used accelerometer data from all available sensors, which are already at 50 Hz, allowing us to employ the data directly without additional preprocessing. For all four datasets, we applied $Z$-score normalization and used a sliding window of 100 samples in length (i.e., two seconds) with 50 samples overlap to generate instances. Furthermore, since each target dataset has different sensor placements, we had to pre-train a separate MuJo model for each one. Specifically, during pre-training our models, we used the virtual sensor positions generated by IMUTube that best match the sensor positions of the respective target dataset. Each of the four models was pre-trained on our FiMAD dataset incorporating all four input modalities: text, video, poses, and virtual accelerometer data.

We conducted our experiments based on the standard experimental methodology in the HAR community, i.e., testing on unseen users and evaluating using the Macro $F_1$-Score, a metric that is well-suited for handling class imbalance. For MM-Fit, we used the training, validation, and cross-subject test sets as described in~\citep{stromback2020mmfit}. Since no fixed splits are provided for MyoGym, MotionSense, and MHEALTH, we created five splits, where we randomly divided subjects into independent training, validation, and test sets for each split.
Additionally, we considered two scenarios: training the classifiers with and without the inclusion of the NULL class. This is relevant because the NULL class is not part of our pre-training dataset but is present in real-world scenarios and is included in all evaluation datasets except for MotionSense.

For classification, we utilized a classifier architecture $C$ comprising our sensor encoder $e_{sensor}$, projection $p_{sensor}$, and a classification head $cl$ consisting of two layers. We contrast three scenarios: In the baseline setting, all weights were initialized randomly, meaning the model lacks any pre-trained knowledge. In the other two scenarios, the encoder and projection head were initialized with weights from our pre-trained model. In one case, the weights of the encoder and projection head were fixed (frozen), whereas in the other case, they were trainable and we fine-tuned them during the classifier's training process. We trained with different proportions of the training data, ranging from one to 100 percent, with the training data being sampled randomly per class to ensure balanced representation. To guarantee the comparability of all results, we trained each classifier $C$ with the same settings:
Training was conducted for a maximum of 200 epochs, incorporating early stopping after 25 epochs without improvement. We employed a weighted cross-entropy classification loss to address class imbalance, given by
\begin{equation}
\mathcal{L}_{CL}\;\; = \;\;-\sum_{i=1}^{n}w_{y_i}y_{i}\log(C(x_i)),
\end{equation}
where $n$ is the number of classes, $y_i$ is a one-hot encoded ground truth activity label and $C(x_i)$ is the predicted label vector of $C$ for input $x_i$, that is, the application of our $cl$ classification layers to obtain $cl(p(e(x_i)))$.
The weight $w_{y_i} = \frac{\max{(M)}}{m_{y_i}}$ for class $y_i$ is calculated by dividing the number of data samples of the largest class by the number of instances in class $y_i$ itself.
To ensure the validity of the results, we trained 20 models independently for each configuration, with the training data being randomly selected anew for each run (for proportions less than 100\%) and the non-pre-trained weights being reinitialized randomly.

\paragraph*{Proxy task comparisons}
We compared our approach with Multi-task~\citep{multitask}, Auto-encoder~\citep{reconstruction}, and simCLR~\citep{simclr}. For a fair comparison, all approaches were trained with the same simulated IMU data and the same encoder architecture as our approach. Regarding other configurations, simCLR uses random rotations as augmentation uniformly sampled in $(-\pi,\pi)$ range. The representation was obtained from our projection layer through two dense layers of size 256, 128, and 128, respectively, the first two being followed by a ReLU activation. It was trained with the NTXent loss using the l2 norm.
For Multi-task, we have selected the following augmentations: noising, scaling, negation, horizontal flipping, permutation, and channel shuffling. Noising consists of adding a value sampled from a normal distribution with a mean of 0.0 and a standard deviation of 0.05, while scaling consists of multiplying by values drawn from this same distribution. Negation, horizontal flipping, and channel shuffling are self-evident. Permutation means randomly splitting the time series into four parts and changing their order. Every task has a head that takes as input the encoder's output and consists of one fully connected layer with 256 size and ReLU activation, followed by another fully connected layer with also 256 size but sigmoid activation. For Auto-encoder, the representation is passed to a convolutional decoder that re-creates the original signal. This decoder consists of first a linear layer to bring the projection to $(32, n\_sensors, \frac{win\_size}{4})$ where $n\_sensors$ is the number of sensors in the dataset and $win\_size$ is the original window size. This is followed by batch norm, ReLU and then three blocks of convolution transposed, $0.05$ dropout, ReLU and upsampling. A kernel size of $(1,11)$ was used and the channels change following the sequence $(32,16,8,orig)$, with $orig$ being the original number of sensor channels. 

To our knowledge, this is the first application of such methods to simulated sensor data, as well as their first use in a multi-sensor scenario. While all those methods could be easily added to our framework, we selected them as a comparison to highlight how our method can learn from extra modalities that are only available during training time, beyond what is present in the simulated sensor data itself (which is the only input for those methods).
\subsection{Multimodal Classification on MM-Fit}
We evaluated the multimodal capabilities of our system using the MM-Fit dataset. Since MM-Fit comes along with videos and poses at 30 Hz, we upsample both modalities to 50 Hz. We extracted HERO CLIP features from the videos following the approach outlined in Section \ref{dataset_construction}. We used the sliding window method described in Section \ref{um-class} to generate instances for pose and video inputs. We also employed the same classifier architecture $C$ but selected the encoders $e_m$ and projections $p_m$ depending on the input modality. Specifically, we trained individual classifier models for the pose and video input modality, using their respective encoders $e_m$ and projections $p_m$. Additionally, we experimented with multimodal inputs by providing all three modalities (sensor, pose, and video) to the model and concatenating the outputs from the respective projections before feeding them into the classifier head $cl$. 
We kept the same training setup as outlined for unimodal training in Section \ref{um-class}, performed training runs with and without the NULL class, and contrasted the same three training scenarios: baseline, pre-trained, and pre-trained with frozen encoder and projection weights. For the pre-trained scenarios, we reused the MuJo model from Section \ref{um-class} that was pre-trained for the MM-Fit target dataset (i.e., allowing left and right wrist sensor placements).

\begin{table*}[!t]
\centering
\resizebox{1.0\textwidth}{!}{
\begin{NiceTabular}{|c|c|c|ccc|ccc|}
\hline
\multirow{3}{*}{\textbf{Dataset}} & \multirow{3}{*}{\shortstack{\textbf{With}\\\textbf{``NULL''}}} & \multirow{3}{*}{\textbf{Method}} & \multicolumn{6}{c}{\textbf{Mean Macro $F_1$-Score}} \\\cline{4-9}
& & & \multicolumn{3}{c}{Train Data: \(2\%\)} & \multicolumn{3}{c}{Train Data: \(100\%\)}\\\cline{4-9}
& &  & Baseline (\trainable) & Pre-trained (\frozen) & Pre-trained (\trainable) & Baseline (\trainable) & Pre-trained (\frozen) & Pre-trained (\trainable)\\
\hline

\multirow{8}{*}{MyoGym} & \multirow{4}{*}{\xmark} & MuJo & \multirow{4}{*}{ \( .242 \pm .011 \)} & \( \mathbf{.317} \pm .013 \) & \( \mathbf{.317} \pm .009 \) & \multirow{4}{*}{\( .459 \pm .012 \)} & \( .469 \pm .006 \) & \( \mathbf{.475} \pm .006 \) \\ 
&  & Auto-encoder &  & \(.065 \pm .016\) & \(.253 \pm .019\)  &  & \(.372 \pm .043\) & \(.435 \pm .024\)  \\
&  & simCLR &  & \( .038 \pm .006 \) & \( .088 \pm .016 \)  &  & \(.081 \pm .022\) & \(.447 \pm .043\)  \\
&  & Multi-task &  & \( .241 \pm .025 \) & \( .278 \pm .029 \) &  &  \(.425 \pm .052\) &  \(.465 \pm .043\)  \\

\cdashline{2-9}

& \multirow{4}{*}{\cmark}  & MuJo & \multirow{4}{*}{\( .184 \pm .007 \)} & \( .229 \pm .010 \) & \( \mathbf{.256} \pm .011 \) & \multirow{4}{*}{\( .353 \pm .009 \)} & \( .376 \pm .007 \) & \( \mathbf{.408} \pm .006 \) \\
&  & Auto-encoder &  & \( .116 \pm .001 \) & \( .204 \pm .002 \)  &  &  \(.242 \pm .028\) &  \(.344 \pm .020 \) \\
&  & simCLR &  & \( .046 \pm .005 \) & \( .094 \pm .015 \)  &  &  \(.102 \pm .018\) &  \(.352 \pm .038\) \\
&  & Multi-task &  & \( .194 \pm .022 \) & \( .233 \pm .002 \)  &  &  \(.339 \pm .047\) &  \(.376 \pm .040\) \\
\hline

\multirow{8}{*}{MHEALTH} & \multirow{4}{*}{\xmark} & MuJo & \multirow{4}{*}{\( .613 \pm .027 \)} & \( .688 \pm .020 \) & \( \mathbf{.700} \pm .026 \) & \multirow{4}{*}{\( .837 \pm .030 \)} & \( .766 \pm .010 \) & \( .839 \pm .018 \) \\
&  & Auto-encoder &  &\( .527 \pm .042 \) & \( .538 \pm .039 \) &  & \( .785 \pm .343 \) & \( .826 \pm .033 \)  \\
&  & simCLR &  &\( .542 \pm .023 \)  & \( .576 \pm .037 \) &  & \( .718 \pm .541 \) & \( \mathbf{.860} \pm .053 \) \\
&  & Multi-task &  & \( .534 \pm .083 \) & \( .626 \pm .099 \) &  & \( .758 \pm .079 \) & \( .758 \pm .079 \) \\

\cdashline{2-9}

& \multirow{4}{*}{\cmark} & MuJo & \multirow{4}{*}{\( .371 \pm .016 \)} & \( .425 \pm .012 \) & \( \mathbf{.428} \pm .021 \) & \multirow{4}{*}{\( .577 \pm .015 \)} & \( .533 \pm .007 \) & \( .582 \pm .013 \) \\
&  & Auto-encoder &  & \( .321 \pm .155 \) & \( .362 \pm .026 \) &  & \( .514 \pm .030 \) & \( .539 \pm .040 \)  \\
&  & simCLR &  & \( .347 \pm .018 \) & \( .386 \pm .034 \) &  & \( .464 \pm .027 \) & \( \mathbf{.602} \pm .053 \) \\
&  & Multi-task &  &\( .313 \pm .032 \) & \( .407 \pm .048 \)  &  & \( .414 \pm .050 \) &  \( .547 \pm .056 \) \\

\hline

\multirow{8}{*}{MM-Fit} & \multirow{4}{*}{\xmark} & MuJo & \multirow{4}{*}{\( .423 \pm .045 \)} & \( .474 \pm .028 \) & \( .492 \pm .027 \) & \multirow{4}{*}{\( .544 \pm .025 \)} & \( .557 \pm .005 \) & \( .580 \pm .014 \) \\
&  & Auto-encoder &  & \( .190 \pm .061 \) & \( .394 \pm .070 \) &  & \( .511 \pm .009 \)  & \( .556 \pm .020 \) \\
&  & simCLR &  & \( .135 \pm .015 \) & \( .337 \pm .036 \) &  & \( .315 \pm .016 \)  & \( .522 \pm .023 \) \\
&  & Multi-task &  &  \( .512 \pm .021 \) & \( \mathbf{.520} \pm .029 \) &  & \( \mathbf{.614} \pm .008 \)  & \( .589 \pm .019 \) \\

\cdashline{2-9}

& \multirow{4}{*}{\cmark} & MuJo & \multirow{4}{*}{\( .257 \pm .023 \)} & \( .237 \pm .022 \) & \( .312 \pm .015 \) & \multirow{4}{*}{\( .389 \pm .028 \)} & \( .368 \pm .006 \) & \( .408 \pm .010 \) \\
&  & Auto-encoder &  & \( .232 \pm .187 \) & \( .243 \pm .037 \) &  & \( .283 \pm .009 \) & \( .382 \pm .011 \)  \\
&  & simCLR &  & \( .112 \pm .009 \) & \( .223 \pm .020 \) &  & \( .202 \pm .011 \) & \( .356 \pm .019 \) \\
&  & Multi-task &  & \( .324 \pm .020 \) & \( \mathbf{.336} \pm .020 \) &  & \( .381 \pm .013 \) & \( \mathbf{.415} \pm .007 \)  \\

\hline

\multirow{4}{*}{MotionSense} & \multirow{4}{*}{\xmark} & MuJo & \multirow{4}{*}{\( .675 \pm .010 \)} & \( .697 \pm .009 \) & \( \mathbf{.734} \pm .011 \) & \multirow{4}{*}{\( .819 \pm .009 \)} & \( .742 \pm .005 \) & \( .845 \pm .004 \) \\
&  & Auto-encoder &  & \( .693 \pm .020 \) & \( .699 \pm .022 \) &  & \( .830 \pm .035 \) & \( \mathbf{.862} \pm .039 \) \\
&  & simCLR &  & \( .640 \pm .385 \) & \( .699 \pm .324 \)  &  & \( .729 \pm .031 \) & \( .850 \pm .048 \)   \\
&  & Multi-task &  & \( .598 \pm .377 \) & \( .691 \pm .445 \) &  & \( .725 \pm .040 \) & \( .834 \pm .055 \) \\

\hline
\end{NiceTabular}
}
\caption{Comparison of classification results on accelerometer data varying the training amount using different methods for pre-training encoders. We also investigate the effect of freezing encoders during fine-tuning. Each experiment was conducted 20 times.}
\label{tab:classification-all-ds}
\end{table*}

\begin{figure*}[!t]
\centering
\includegraphics[width=0.9\textwidth]{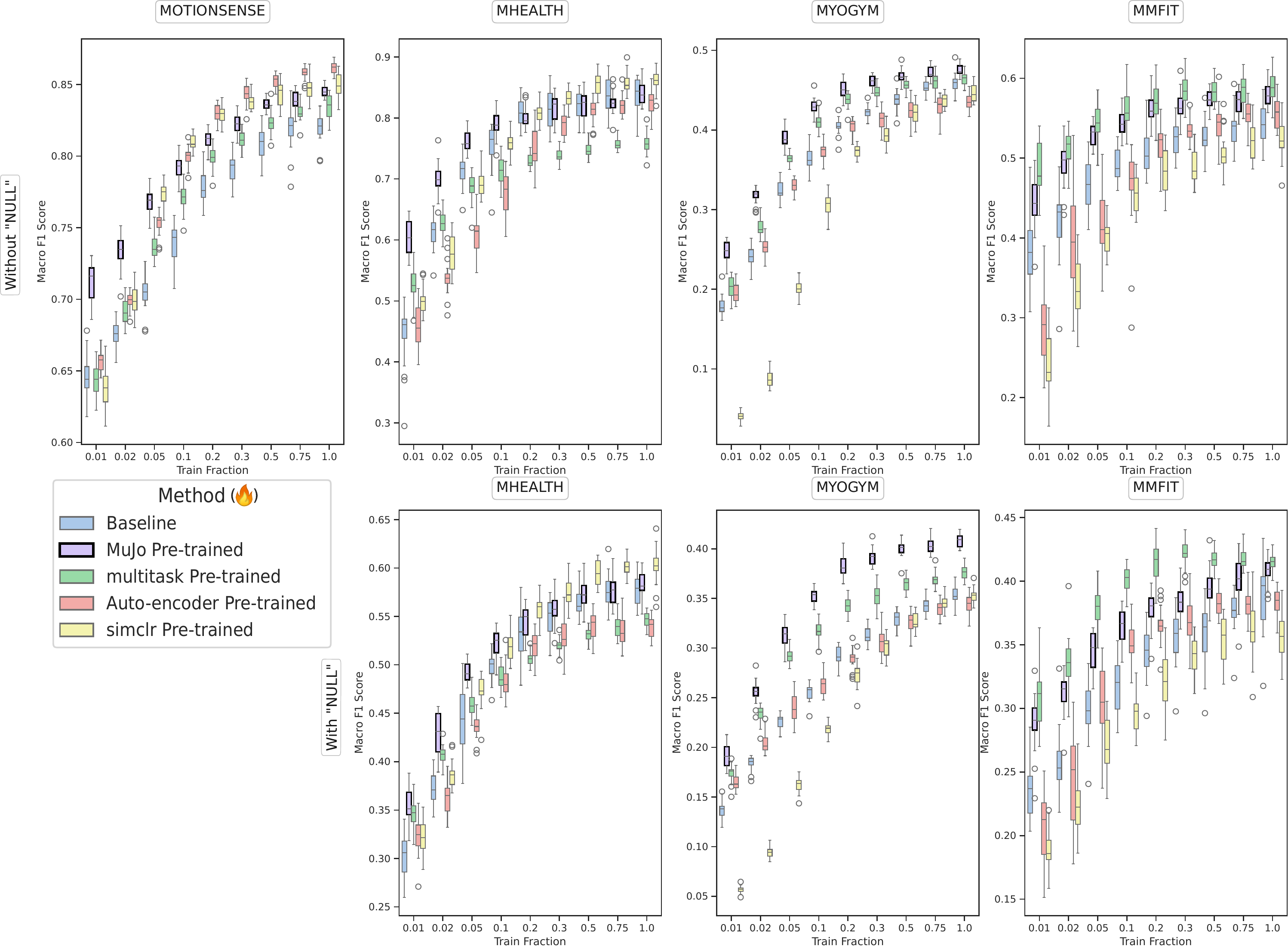}
\caption{\centering Classification performance across various methods and training data fractions, where each boxplot represents the results of 20 runs. The comparison includes the baseline model and models with a pre-trained encoder and projection with trainable weights on accelerometer data for all evaluated datasets.}
\label{fig:comp_all_bps}
\end{figure*}

\subsection{Zero-Shot}
Beyond classification, we evaluated the ability of our pre-trained MuJo model to generalize to unseen datasets without additional fine-tuning. We first calculated the ADA text embeddings $v^l_{text}$ for all labels $l$. For the NULL class, which doesn't correspond to a specific activity, we generated its embedding by taking the mean of the embeddings of all labels. For a given data sample $v_m$ of input modality $m$, we then chose labels based on the closest similarity score $sim(v^l_{text}, v_m)$ by calculating $\underset{l}{\mathrm{argmin}} \, sim(v^l_{text}, v_m)$ for all labels $l$.
We contrasted the outcomes of unimodal training—where MuJo models were individually trained to align sensor-text, pose-text, or video-text—with those of multimodal training, where we tasked a single model to align all four modalities concurrently. We conducted experiments on all four datasets, both with and without the NULL class, and calculated the accuracy and Macro $F1$-Score.

\section{Results and Discussion}
\subsection{Unimodal Classification on Real-World Accelerometer Data}

Table \ref{tab:classification-all-ds} shows that for our method the pre-trained models with trainable weights consistently outperform the baseline with respect to the Macro $F_1$-Score on all datasets, regardless of whether 2\% or 100\% of the training data are used. These findings indicate that multimodal pre-training via contrastive learning on virtually generated accelerometer data can effectively improve HAR performance on real-world sensor data. In comparison with other approaches, we can see that our method is more data-efficient, as we outperform other approaches in almost all cases when 2\% of the labeled training data are available. When 100\% of the training data are available, every method is dominant in a different dataset, but ours consistently improves on the baseline unlike other approaches. Moreover, our method presents a more consistent performance when freezing the encoder, showing the quality of the learned representation.

In general, including the NULL class results in lower scores, but our trainable models still achieve higher scores than the baseline, even if this class was not part of our pre-training. Only for the MHEALTH dataset, when utilizing 100\% of the data, the performance difference becomes marginal, with scores of 0.839 versus 0.837 without the NULL class and 0.582 versus 0.577 with the NULL class.
However, when labeled data are limited, we perform significantly better than the baseline. For example, when utilizing only 2\% of the train data, even the pre-trained classifiers with frozen encoder and projection layers outperform the baseline in all cases except for the MM-Fit dataset when the NULL class is included. 

In Figure \ref{fig:comp_all_bps}, we compare MuJo with the baseline. Moreover, we compare it with three other commonly used pre-training methods, namely Multi-task~\citep{multitask}, Auto-encoder~\citep{reconstruction}, and simCLR~\citep{simclr}, which we also pre-trained on our FiMAD dataset. The boxplots show the Macro $F_1$-Scores of the experiments conducted with different training data fractions ranging from 1\% to 100\%, all with trainable encoders. Our method constantly achieves higher scores than the baseline for all training fractions on all datasets, which demonstrates the effectiveness of our pre-trained representations for accelerometer-based activity recognition. Moreover, it outperforms all other pre-training methods on the MyoGym dataset. Additionally, we achieve better performance for training fractions up to 10\% in all dataset except MM-Fit, where only Multi-task is better. For higher train fractions, simCLR slightly outperforms us on MHEALTH, and both simCLR and Auto-encoder surpass us on MotionSense.

\subsection{Multimodal Classification on MM-Fit}
\begin{figure*}[!t]
\centering
\includegraphics[width=0.75\textwidth]{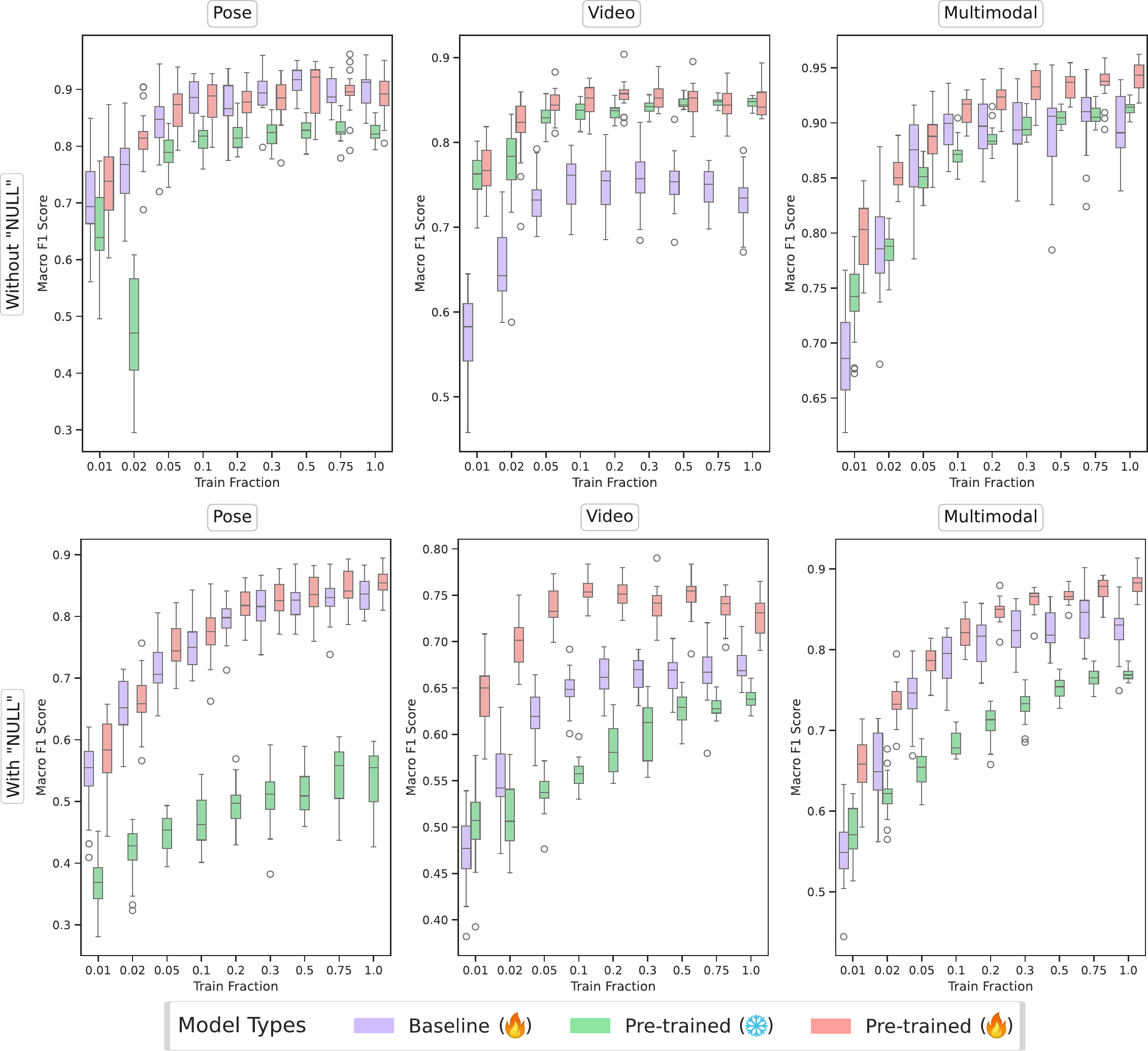}
\caption{Classification performance across various training data fractions on MM-Fit, where each boxplot represents the results of 20 runs. The baseline model is compared to models with a pre-trained encoder and projection (both with frozen and trainable weights) on all input modalities (sensor, pose, video, and multimodal).}
\label{fig:boxplots-mmfit-all-mods}
\end{figure*}

Table \ref{tab:classification-mmfit} presents the results of our experiments conducted for pose, video, and multimodal (concatenation of sensor, pose, and video) inputs, using both 2\% and 100\% of the training data, and with and without the inclusion of the NULL class. While the classifiers with our frozen pre-trained MuJo encoders and projections often perform worse than the baseline, our trainable models outperform the baseline in all cases except for the pose input without the NULL class when using the full training data. In general, for the pose input, the performance difference between these two training scenarios is only marginal, as evident from Figure \ref{fig:boxplots-mmfit-all-mods}. This discrepancy may be attributed to differences between the poses generated by IMUTube and that extracted by MM-Fit. Although both IMUTube and MM-Fit utilize VidePose3D for pose extraction, IMUTube incorporates extensive additional pre- and post-processing steps. This hypothesis is supported by the results for the video modality, where our trainable models outperform the baseline in all scenarios. Since the video embeddings, which we used as inputs, were extracted in the same manner for both our FiMAD dataset and MM-Fit, there is no such inconsistency, unlike the poses.

Leveraging multimodal inputs achieves the highest performance, suggesting that HAR benefits from the diverse perspectives of multiple modalities. In this multimodal scenario, our method, compared to the baseline, improves the Macro $F_1$-Score from 0.824 to 0.882 (with NULL) and from 0.897 to 0.942 (without NULL) when trained on the full dataset. Furthermore, with only 2\% of the training data available, we improve performance from 0.656 to 0.734 (with NULL) and from 0.790 to 0.855 (without NULL). These findings show that the pre-trained classifiers significantly outperform the baseline and learn more efficiently with a limited amount of training data across multiple input modalities.

\subsection{Zero-Shot}
Table \ref{tab:zeroshot} demonstrates zero-shot results of MuJo on all four datasets. For MM-Fit and MotionSense, multimodal training outperforms unimodal training, achieving consistently higher top-1, top-3, and top-5 accuracy and Macro $F_1$-Scores. The MHEALTH dataset shows minimal differences between both training methods. 
However, for the MyoGym dataset, which contains 30 gym exercises where all but two require equipment like machines, barbells, and more, multimodal training performs worse. The reason for this might be that our pre-training was conducted on YouTube videos of fitness trainers primarily performing HIIT workouts, at home and not in the gym, focusing on bodyweight exercises. This domain shift may explain the decrease in performance from unimodal to multimodal training on MyoGym: since the video features in our pre-training dataset lack representations of equipment-based exercises, they provide no valuable information for such types of exercises. Thus, the multimodal model might struggle to recognize activities that involve gym equipment. Consequently, unimodal training performs slightly better on this complex dataset, suggesting that target domain differences significantly impact performance.

The overall low scores are mainly due to data divergences between our pre-training dataset FiMAD and the evaluation datasets, particularly for pose (MM-Fit) and sensor data. Since we pre-trained with virtual sensor data and poses extracted by IMUTube, our models struggle when applied to real sensor datasets and variations in pose representation (MM-Fit), leading to decreased performance. Video features perform the best among all modalities because they were pre-extracted with HERO CLIP in the same way for both datasets, FiMAD and MM-Fit. Further, all our models were not pre-trained with the NULL class, which is represented as the mean of all target class embeddings. Introducing this unknown class during evaluation increases classification complexity and leads to an additional drop in performance.

\begin{table*}[t]
\centering
\resizebox{1.0\textwidth}{!}{
\begin{NiceTabular}{|c|c|ccc|ccc|}

\hline
\multirow{3}{*}{\shortstack{\textbf{With}\\\textbf{``NULL''}}} & \multirow{3}{*}{\textbf{Input Modality}} & \multicolumn{6}{c}{\textbf{Mean Macro $F_1$-Score}} \\\cline{3-8}
&& \multicolumn{3}{c}{Train Data: \(2\%\)} & \multicolumn{3}{c}{Train Data: \(100\%\)}\\\cline{3-8}
&& Baseline (\trainable) & Pre-trained (\frozen) & Pre-trained (\trainable) & Baseline (\trainable) & Pre-trained (\frozen) & Pre-trained (\trainable)\\
\hline
\parbox[t]{2mm}{\multirow{3}{*}{\xmark}} & Pose & \( .759 \pm .063 \) & \( .470 \pm .092 \) & \( \mathbf{.813} \pm .049 \) &\( \mathbf{.900} \pm .033 \) & \( .825 \pm .017 \) & \( .887 \pm .035 \) \\
 & Video & \( .655 \pm .039 \) & \( .772 \pm .054 \) & \( \mathbf{.818} \pm .037 \) &\( .732 \pm .029 \) & \( .847 \pm .006 \) & \( \mathbf{.848} \pm .018 \) \\
 & Multimodal & \( .790 \pm .044 \) & \( .784 \pm .017 \) & \( \mathbf{.855} \pm .016 \) &\( .897 \pm .028 \) & \( .913 \pm .005 \) & \( \mathbf{.942} \pm .012 \) \\
\hline
\parbox[t]{2mm}{\multirow{3}{*}{\cmark}} & Pose & \( .650 \pm .049 \) & \( .418 \pm .043 \) & \( \mathbf{.660} \pm .044 \) &\( .834 \pm .028 \) & \( .533 \pm .051 \) & \( \mathbf{.855} \pm .021 \) \\
 & Video & \( .551 \pm .038 \) & \( .509 \pm .035 \) & \( \mathbf{.699} \pm .025 \) &\( .675 \pm .018 \) & \( .639 \pm .011 \) & \( \mathbf{.729} \pm .021 \) \\
 & Multimodal & \( .656 \pm .043 \) & \( .619 \pm .026 \) & \( \mathbf{.734} \pm .024 \) &\( .824 \pm .027 \) & \( .770 \pm .007 \) & \( \mathbf{.882} \pm .015 \) \\
\hline

\end{NiceTabular}}
\caption{Comparison of classification results for pose, video, and multimodal inputs evaluated on the MM-Fit test dataset when utilizing 2\% and 100\% of the training data. Each experiment was conducted 20 times.}
\label{tab:classification-mmfit}
\end{table*}

\begin{table*}[!t]
\centering
\resizebox{1.0\textwidth}{!}{
\begin{NiceTabular}{|c|c|c|ccc|ccc|ccc|ccc|}
\hline
\multirow{4}{*}{\textbf{Dataset}} & \multirow{4}{*}{\shortstack{\textbf{With}\\\textbf{``NULL''}}} & \multirow{4}{*}{\textbf{Input Modality}} & \multicolumn{12}{c|}{\textbf{Metrics}} \\\cline{4-15}
&&& \multicolumn{6}{c|}{\textbf{Accuracy}} & \multicolumn{6}{c|}{\textbf{Macro $F_1$-Score}} \\\cline{4-15}
&&& \multicolumn{3}{c|}{Unimodal Training} & \multicolumn{3}{c|}{Multimodal Training} & \multicolumn{3}{c|}{Unimodal Training} & \multicolumn{3}{c|}{Multimodal Training} \\\cline{4-15}
&&& Top-1 & Top-3 & Top-5 & Top-1 & Top-3 & Top-5 & Top-1 & Top-3 & Top-5 & Top-1 & Top-3 & Top-5 \\
\hline
\multirow{6}{*}{MM-Fit} & \multirow{3}{*}{\xmark} & Sensor & .112 & .310 & .491 & .169 & .378 & .575 & .084 & .274 & .461 & .114 & .341 & .536 \\
 &  & Pose & .159 & .369 & .587 & .196 & .561 & .671 & .027 & .233 & .443 & .061 & .420 & .574 \\
 &  & Video & .393 & .565 & .652 & .501 & .660 & .733 & .285 & .494 & .611 & .427 & .616 & .712 \\
\cline{2-15}
 & \multirow{3}{*}{\cmark} & Sensor & .027 & .396 & .862 & .064 & .559 & .879 & .027 & .165 & .400 & .036 & .227 & .498 \\
 &  & Pose & .035 & .080 & .127 & .044 & .282 & .409 & .006 & .188 & .370 & .015 & .301 & .475 \\
 &  & Video & .073 & .113 & .454 & .111 & .477 & .920 & .107 & .240 & .395 & .201 & .417 & .680 \\
\hline
\multirow{2}{*}{MHEALTH} & \xmark & Sensor & .117 & .317 & .515 & .100 & .335 & .536 & .093 & .268 & .454 & .081 & .273 & .484 \\
\cline{2-15}
 & \cmark & Sensor & .161 & .584 & .817 & .164 & .638 & .829 & .054 & .213 & .418 & .052 & .203 & .417 \\
\hline
\multirow{2}{*}{MyoGym} & \xmark & Sensor & .083 & .161 & .235 & .052 & .122 & .172 & .054 & .117 & .190 & .032 & .093 & .135 \\
\cline{2-15}
 & \cmark & Sensor & .021 & .046 & .147 & .009 & .026 & .070 & .021 & .054 & .113 & .005 & .041 & .078 \\
\hline
MotionSense & \xmark & Sensor & .143 & .428 & .765 & .146 & .481 & .867 & .076 & .353 & .745 & .104 & .453 & .850 \\
\hline
\end{NiceTabular}}
\caption{Consolidated zero-shot results contrasting uni- and multimodal training of MuJo on FiMAD.}
\label{tab:zeroshot}
\end{table*}

\begin{table*}[!t]
\centering
\resizebox{1.0\textwidth}{!}{
\begin{NiceTabular}{|l|rrrrrr|}
\hline
\textbf{Modality} & \textbf{Encoder FLOPs} & \textbf{Encoder Params} & \textbf{Projection FLOPs} & \textbf{Projection Params} & \textbf{Total FLOPs} & \textbf{Total Params} \\
\hline
Text   & 2.3601 M & 1.1804 M & 8.5286 M & 4.2662 M & 10.8887 M & 5.4467 M \\
Pose   & 38.1508 M & 2.3296 M & 9.1840 M & 4.5939 M & 47.3348 M & 6.9235 M \\
Video  & 655.8720 K & 328.1920 K & 7.2179 M & 3.6109 M & 7.8738 M & 3.9391 M \\
Sensor (MyoGym) & 2.1521 M & 619.5140 K & 7.8733 M & 3.9386 M & 10.0254 M & 4.5581 M \\
Sensor (MHEALTH) & 2.1917 M & 619.7120 K & 7.8733 M & 3.9386 M & 10.0650 M & 4.5583 M \\
Sensor (MotionSense) & 2.1521 M & 619.5140 K & 7.8733 M & 3.9386 M & 10.0254 M & 4.5581 M \\
Sensor (MM-Fit) & 2.1719 M & 619.6130 K & 7.8733 M & 3.9386 M & 10.0452 M & 4.5582 M \\
\hline
\end{NiceTabular}}
\caption{FLOPs (forward pass) and parameters for encoder and projection by modality.}
\label{tab:flops_and_params}
\end{table*}

\subsection{Computational Requirements}
Table \ref{tab:flops_and_params} shows the computational requirements of our models. For each modality, we calculated the number of parameters for the encoders, projections, and their total. The table shows that all models are compact, with parameters ranging between 3.94M and 6.92M parameters, resulting in model sizes of only a few MB. Additionally, we calculated the theoretical amount of floating-point operations (FLOPs) required for a forward pass with a batch size of 1. Except for the pose encoder, all models are highly efficient, with total FLOPs below 11M. The text and video encoders are particularly compact because they operate on feature vectors pre-extracted with the ADA model for text and the HERO CLIP model for video, rather than operating on raw text and video data. For the sensor encoders, the number of sensors (one for MHEALTH, two for MM-Fit, and three for MyoGym and MotionSense) has only a minor impact on size and computational complexity. The pose encoder requires the highest number of FLOPs (38.15M), due to its complex model structure and input dimensions of (3, 17, 100), where 3 represents the number of channels, 17 the number of joints, and 100 the window size.

In addition to the inference performance of the encoders and projections, we also analyzed the time required to pre-train our MuJo model, which is notably efficient. For instance, training MuJo with all four input modalities (text, sensor, video, and pose) on a A100 GPU requires less than 1 hour and uses less than 3GB of VRAM.

\section{Limitations}
This study has two main limitations. First, it focuses exclusively on the fitness domain for HAR. While this specialization enables the development of a targeted and optimized approach, it restricts the generalizability of our findings to other domains, such as healthcare and general daily activity recognition. Second, the feature extraction process for each modality relies on pre-trained models. Although this leverages state-of-the-art methods and reduces computational requirements during training, it also introduces dependencies on the quality and biases of these models. For instance, pre-trained pose estimation models may perform poorly on unusual or occluded postures, and sensor-based models might not generalize well across different devices or user populations. These limitations highlight the need for more domain-agnostic and robust solutions in future work. It is also important to highlight the limited zero-shot performance when the target classes diverge too much, be it by the presence of gym equipment or of classes where the subject is not performing any exercise (null class).  

\section{Conclusion and Future Work}
In this study, we introduced FiMAD, a multimodal dataset that contains over 78 hours of recordings. Our method, MuJo, is a novel pre-training approach that learns a joint feature space integrating video, pose, simulated accelerometer, and textual data. Our approach provides gains in real datasets such as MM-Fit, MyoGym, MotionSense, and MHEALTH, showing that our pre-training method captures a meaningful representation for HAR. Despite being trained solely on virtual IMU data, it enhances performance on real accelerometer data, especially when less data were used for training. Even more, our approach offers the flexibility to be applied not only to sensor inputs but also to video and pose inputs. Finally, although our zero-shot results are not outstanding, they demonstrate that multimodal training can enhance performance compared to unimodal training, if the target domain does not differ too much from our pre-training domain.

While our study presents a robust and innovative approach to HAR by integrating multiple data modalities into a joint feature space, there are several avenues for future research and enhancement. One potential improvement is to fine-tune the text and vision models during pre-training, rather than using pre-calculated text and video embeddings. Additionally, we plan to conduct a more detailed evaluation in the future. This will include extending our pre-training approach to other HAR data sources with real sensor data, as well as training on the target dataset itself, rather than relying solely on models pre-trained with virtual accelerometer data. Furthermore, we aim to enhance the handling of the NULL class, which can be achieved by using a large language model to label the chapters of FiMAD that do not contain relevant activities as NULL. This would provide a clear representation for the NULL class in our dataset. Future work also includes incorporating other proxy tasks in our framework during pre-training as well as scaling the networks. In that setup, we aim to train a single MuJo model instead of training a separate one for each target dataset when the sensor positions differ.

\bibliographystyle{unsrtnat}
\bibliography{mybibfile.bib}

\begin{thebibliography}{52}
\providecommand{\natexlab}[1]{#1}
\providecommand{\url}[1]{\texttt{#1}}
\expandafter\ifx\csname urlstyle\endcsname\relax
  \providecommand{\doi}[1]{doi: #1}\else
  \providecommand{\doi}{doi: \begingroup \urlstyle{rm}\Url}\fi

\bibitem[Tan et~al.(2021)Tan, Beheshti, Binnie, Davey, Caneiro, Kent, Smith, O’Sullivan, and Campbell]{tan2021human}
Jay-Shian Tan, Behrouz~Khabbaz Beheshti, Tara Binnie, Paul Davey, J.~P. Caneiro, Peter Kent, Anne Smith, Peter O’Sullivan, and Amity Campbell.
\newblock Human activity recognition for people with knee osteoarthritis—a proof-of-concept.
\newblock \emph{Sensors}, 21\penalty0 (10):\penalty0 3381, 2021.

\bibitem[Host and Iva{\v{s}}i{\'c}-Kos(2022)]{host2022overview}
Kristina Host and Marina Iva{\v{s}}i{\'c}-Kos.
\newblock An overview of human action recognition in sports based on computer vision.
\newblock \emph{Heliyon}, page e09633, 2022.

\bibitem[Nadeem et~al.(2020)Nadeem, Jalal, and Kim]{nadeem2020accurate}
Amir Nadeem, Ahmad Jalal, and Kibum Kim.
\newblock Accurate physical activity recognition using multidimensional features and markov model for smart health fitness.
\newblock \emph{Symmetry}, 12\penalty0 (11):\penalty0 1766, 2020.

\bibitem[Sunil et~al.(2021)Sunil, Sheth, Shreyas, et~al.]{sunil2021usual}
Ajeet Sunil, Manav~Hiren Sheth, E~Shreyas, et~al.
\newblock Usual and unusual human activity recognition in video using deep learning and artificial intelligence for security applications.
\newblock In \emph{2021 Fourth International Conference on Electrical, Computer and Communication Technologies (ICECCT)}, pages 1--6. IEEE, 2021.

\bibitem[Radford et~al.(2021)]{radford2021learning}
Alec Radford et~al.
\newblock Learning transferable visual models from natural language supervision.
\newblock In \emph{International conference on machine learning}, pages 8748--8763. PMLR, 2021.

\bibitem[Achiam et~al.(2023)Achiam, Adler, Agarwal, Ahmad, Akkaya, Aleman, Almeida, Altenschmidt, Altman, Anadkat, et~al.]{achiam2023gpt}
Josh Achiam, Steven Adler, Sandhini Agarwal, Lama Ahmad, Ilge Akkaya, Florencia~Leoni Aleman, Diogo Almeida, Janko Altenschmidt, Sam Altman, Shyamal Anadkat, et~al.
\newblock Gpt-4 technical report.
\newblock \emph{arXiv preprint arXiv:2303.08774}, 2023.

\bibitem[Kelly et~al.(2025)Kelly, Longjohn, and Nottingham]{kelly2025uci}
Markelle Kelly, Rachel Longjohn, and Kolby Nottingham.
\newblock The uci machine learning repository, 2025.
\newblock URL \url{https://archive.ics.uci.edu}.
\newblock Accessed: 2025-01-29.

\bibitem[Fortes~Rey et~al.(2021)Fortes~Rey, Garewal, and Lukowicz]{rey2021translating}
Vitor Fortes~Rey, Kamalveer~Kaur Garewal, and Paul Lukowicz.
\newblock Translating videos into synthetic training data for wearable sensor-based activity recognition systems using residual deep convolutional networks.
\newblock \emph{Applied Sciences}, 11\penalty0 (7), 2021.
\newblock ISSN 2076-3417.
\newblock \doi{10.3390/app11073094}.

\bibitem[Carreira et~al.(2019)Carreira, Noland, Hillier, and Zisserman]{carreira2019short}
Joao Carreira, Eric Noland, Chloe Hillier, and Andrew Zisserman.
\newblock A short note on the kinetics-700 human action dataset.
\newblock \emph{arXiv preprint arXiv:1907.06987}, 2019.

\bibitem[Chan et~al.(2024)Chan, Yuan, Tong, Acquah, Schonfeldt, Gershuny, and Doherty]{chan2024capture24}
Shing Chan, Hang Yuan, Catherine Tong, Aidan Acquah, Abram Schonfeldt, Jonathan Gershuny, and Aiden Doherty.
\newblock Capture-24: A large dataset of wrist-worn activity tracker data collected in the wild for human activity recognition, 2024.

\bibitem[Kwon et~al.(2021{\natexlab{a}})Kwon, Wang, Abowd, and Pl\"{o}tz]{kwon2021imutube2}
Hyeokhyen Kwon, Bingyao Wang, Gregory~D. Abowd, and Thomas Pl\"{o}tz.
\newblock Approaching the real-world: Supporting activity recognition training with virtual imu data.
\newblock \emph{Proc. ACM Interact. Mob. Wearable Ubiquitous Technol.}, 5\penalty0 (3), sep 2021{\natexlab{a}}.

\bibitem[Cao et~al.(2017)Cao, Simon, Wei, and Sheikh]{cao2017realtime}
Zhe Cao, Tomas Simon, Shih-En Wei, and Yaser Sheikh.
\newblock Realtime multi-person 2d pose estimation using part affinity fields.
\newblock In \emph{Proceedings of the IEEE conference on computer vision and pattern recognition}, pages 7291--7299, 2017.

\bibitem[Pavllo et~al.(2019)Pavllo, Feichtenhofer, Grangier, and Auli]{pavllo20193d}
Dario Pavllo, Christoph Feichtenhofer, David Grangier, and Michael Auli.
\newblock 3d human pose estimation in video with temporal convolutions and semi-supervised training.
\newblock In \emph{Proceedings of the IEEE/CVF conference on computer vision and pattern recognition}, pages 7753--7762, 2019.

\bibitem[Str\"{o}mb\"{a}ck et~al.(2020)Str\"{o}mb\"{a}ck, Huang, and Radu]{stromback2020mmfit}
David Str\"{o}mb\"{a}ck, Sangxia Huang, and Valentin Radu.
\newblock Mm-fit: Multimodal deep learning for automatic exercise logging across sensing devices.
\newblock \emph{Proc. ACM Interact. Mob. Wearable Ubiquitous Technol.}, 4\penalty0 (4), dec 2020.

\bibitem[Koskim{\"a}ki et~al.(2017)Koskim{\"a}ki, Siirtola, and R{\"o}ning]{koskimaki2017myogym}
Heli Koskim{\"a}ki, Pekka Siirtola, and Juha R{\"o}ning.
\newblock Myogym: introducing an open gym data set for activity recognition collected using myo armband.
\newblock In \emph{Proceedings of the 2017 ACM international joint conference on pervasive and ubiquitous computing and proceedings of the 2017 ACM international symposium on wearable computers}, pages 537--546, 2017.

\bibitem[Malekzadeh et~al.(2018)Malekzadeh, Clegg, Cavallaro, and Haddadi]{malekzadeh2018motionsense}
Mohammad Malekzadeh, Richard~G Clegg, Andrea Cavallaro, and Hamed Haddadi.
\newblock Protecting sensory data against sensitive inferences.
\newblock In \emph{Proceedings of the 1st Workshop on Privacy by Design in Distributed Systems}, pages 1--6, 2018.

\bibitem[Banos et~al.(2015)Banos, Villalonga, Garcia, Saez, Damas, Holgado-Terriza, Lee, Pomares, and Rojas]{banos2015mhealth}
Oresti Banos, Claudia Villalonga, Rafael Garcia, Alejandro Saez, Miguel Damas, Juan~A Holgado-Terriza, Sungyong Lee, Hector Pomares, and Ignacio Rojas.
\newblock Design, implementation and validation of a novel open framework for agile development of mobile health applications.
\newblock \emph{Biomedical engineering online}, 14:\penalty0 1--20, 2015.

\bibitem[Verma et~al.(2020)Verma, Kumawat, Nakashima, and Raman]{verma2020yoga}
Manisha Verma, Sudhakar Kumawat, Yuta Nakashima, and Shanmuganathan Raman.
\newblock Yoga-82: a new dataset for fine-grained classification of human poses.
\newblock In \emph{Proceedings of the IEEE/CVF conference on computer vision and pattern recognition workshops}, pages 1038--1039, 2020.

\bibitem[Shao et~al.(2020)Shao, Zhao, Dai, and Lin]{shao2020finegym}
Dian Shao, Yue Zhao, Bo~Dai, and Dahua Lin.
\newblock Finegym: A hierarchical video dataset for fine-grained action understanding.
\newblock In \emph{Proceedings of the IEEE/CVF conference on computer vision and pattern recognition}, pages 2616--2625, 2020.

\bibitem[Fieraru et~al.(2021)Fieraru, Zanfir, Pirlea, Olaru, and Sminchisescu]{fieraru2021aifit}
Mihai Fieraru, Mihai Zanfir, Silviu~Cristian Pirlea, Vlad Olaru, and Cristian Sminchisescu.
\newblock Aifit: Automatic 3d human-interpretable feedback models for fitness training.
\newblock In \emph{Proceedings of the IEEE/CVF conference on computer vision and pattern recognition}, pages 9919--9928, 2021.

\bibitem[Kwon et~al.(2021{\natexlab{b}})Kwon, Abowd, and Pl{\"o}tz]{kwon2021complex}
Hyeokhyen Kwon, Gregory~D Abowd, and Thomas Pl{\"o}tz.
\newblock Complex deep neural networks from large scale virtual imu data for effective human activity recognition using wearables.
\newblock \emph{Sensors}, 21\penalty0 (24):\penalty0 8337, 2021{\natexlab{b}}.

\bibitem[Thukral et~al.(2023)Thukral, Haresamudram, and Ploetz]{thukral2023cross}
Megha Thukral, Harish Haresamudram, and Thomas Ploetz.
\newblock Cross-domain har: Few shot transfer learning for human activity recognition.
\newblock \emph{ACM Transactions on Intelligent Systems and Technology}, 2023.

\bibitem[Kamboj and Do(2024)]{kamboj2024survey}
Abhi Kamboj and Minh Do.
\newblock A survey of imu based cross-modal transfer learning in human activity recognition.
\newblock \emph{arXiv preprint arXiv:2403.15444}, 2024.

\bibitem[Kwapisz et~al.(2011)Kwapisz, Weiss, and Moore]{kwapisz2011activity}
Jennifer~R Kwapisz, Gary~M Weiss, and Samuel~A Moore.
\newblock Activity recognition using cell phone accelerometers.
\newblock \emph{ACM SigKDD Explorations Newsletter}, 12\penalty0 (2):\penalty0 74--82, 2011.

\bibitem[Ord{\'o}{\~n}ez and Roggen(2016)]{ordonez2016deep}
Francisco~Javier Ord{\'o}{\~n}ez and Daniel Roggen.
\newblock Deep convolutional and lstm recurrent neural networks for multimodal wearable activity recognition.
\newblock \emph{Sensors}, 16\penalty0 (1):\penalty0 115, 2016.

\bibitem[Mekruksavanich and Jitpattanakul(2022)]{mekruksavanich2022multimodal}
Sakorn Mekruksavanich and Anuchit Jitpattanakul.
\newblock Multimodal wearable sensing for sport-related activity recognition using deep learning networks.
\newblock \emph{Journal of Advances in Information Technology}, 2022.

\bibitem[Ijaz et~al.(2022)Ijaz, Diaz, and Chen]{ijaz2022multimodal}
Momal Ijaz, Renato Diaz, and Chen Chen.
\newblock Multimodal transformer for nursing activity recognition.
\newblock In \emph{Proceedings of the IEEE/CVF Conference on Computer Vision and Pattern Recognition}, pages 2065--2074, 2022.

\bibitem[Duhme et~al.(2022)Duhme, Memmesheimer, and Paulus]{duhme2022fusion}
Michael Duhme, Raphael Memmesheimer, and Dietrich Paulus.
\newblock Fusion-gcn: Multimodal action recognition using graph convolutional networks.
\newblock In \emph{Pattern Recognition: 43rd DAGM German Conference, DAGM GCPR 2021, Bonn, Germany, September 28--October 1, 2021, Proceedings}, pages 265--281. Springer, 2022.

\bibitem[Plananamente et~al.(2022)Plananamente, Plizzari, and Caputo]{plananamente2022test}
Mirco Plananamente, Chiara Plizzari, and Barbara Caputo.
\newblock Test-time adaptation for egocentric action recognition.
\newblock In \emph{Image Analysis and Processing--ICIAP 2022: 21st International Conference, Lecce, Italy, May 23--27, 2022, Proceedings, Part III}, pages 206--218. Springer, 2022.

\bibitem[Islam et~al.(2022)Islam, Nooruddin, Karray, and Muhammad]{islam2022human}
Md~Milon Islam, Sheikh Nooruddin, Fakhri Karray, and Ghulam Muhammad.
\newblock Human activity recognition using tools of convolutional neural networks: A state of the art review, data sets, challenges, and future prospects.
\newblock \emph{Computers in Biology and Medicine}, page 106060, 2022.

\bibitem[Seo et~al.(2021)Seo, Nagrani, and Schmid]{seo2021look}
Paul~Hongsuck Seo, Arsha Nagrani, and Cordelia Schmid.
\newblock Look before you speak: Visually contextualized utterances.
\newblock In \emph{Proceedings of the IEEE/CVF Conference on Computer Vision and Pattern Recognition}, pages 16877--16887, 2021.

\bibitem[Sun et~al.(2019)Sun, Myers, Vondrick, Murphy, and Schmid]{sun2019videobert}
Chen Sun, Austin Myers, Carl Vondrick, Kevin Murphy, and Cordelia Schmid.
\newblock Videobert: A joint model for video and language representation learning.
\newblock In \emph{Proceedings of the IEEE/CVF international conference on computer vision}, pages 7464--7473, 2019.

\bibitem[Li et~al.(2020)Li, Chen, Cheng, Gan, Yu, and Liu]{li2020hero}
Linjie Li, Yen-Chun Chen, Yu~Cheng, Zhe Gan, Licheng Yu, and Jingjing Liu.
\newblock Hero: Hierarchical encoder for video+ language omni-representation pre-training.
\newblock \emph{arXiv preprint arXiv:2005.00200}, 2020.

\bibitem[Miech et~al.(2020)Miech, Alayrac, Smaira, Laptev, Sivic, and Zisserman]{miech2020end}
Antoine Miech, Jean-Baptiste Alayrac, Lucas Smaira, Ivan Laptev, Josef Sivic, and Andrew Zisserman.
\newblock End-to-end learning of visual representations from uncurated instructional videos.
\newblock In \emph{Proceedings of the IEEE/CVF Conference on Computer Vision and Pattern Recognition}, pages 9879--9889, 2020.

\bibitem[Hu et~al.(2018)Hu, Zheng, Ma, Wang, Lai, and Zhang]{hu2018early}
Jian-Fang Hu, Wei-Shi Zheng, Lianyang Ma, Gang Wang, Jianhuang Lai, and Jianguo Zhang.
\newblock Early action prediction by soft regression.
\newblock \emph{IEEE transactions on pattern analysis and machine intelligence}, 41\penalty0 (11):\penalty0 2568--2583, 2018.

\bibitem[Zadeh et~al.(2017)Zadeh, Chen, Poria, Cambria, and Morency]{zadeh2017tensor}
Amir Zadeh, Minghai Chen, Soujanya Poria, Erik Cambria, and Louis-Philippe Morency.
\newblock Tensor fusion network for multimodal sentiment analysis.
\newblock \emph{arXiv preprint arXiv:1707.07250}, 2017.

\bibitem[Lu et~al.(2019)Lu, Batra, Parikh, and Lee]{lu2019vilbert}
Jiasen Lu, Dhruv Batra, Devi Parikh, and Stefan Lee.
\newblock Vilbert: Pretraining task-agnostic visiolinguistic representations for vision-and-language tasks.
\newblock \emph{Advances in neural information processing systems}, 32, 2019.

\bibitem[Saeed et~al.(2019)Saeed, Ozcelebi, and Lukkien]{multitask}
Aaqib Saeed, Tanir Ozcelebi, and Johan Lukkien.
\newblock Multi-task self-supervised learning for human activity detection.
\newblock \emph{Proceedings of the {ACM} on Interactive, Mobile, Wearable and Ubiquitous Technologies}, 3\penalty0 (2):\penalty0 1--30, jun 2019.

\bibitem[Haresamudram et~al.(2019)Haresamudram, Anderson, and Pl\"{o}tz]{reconstruction}
Harish Haresamudram, David~V. Anderson, and Thomas Pl\"{o}tz.
\newblock On the role of features in human activity recognition.
\newblock In \emph{Proceedings of the 2019 ACM International Symposium on Wearable Computers}, ISWC '19, page 78–88, New York, NY, USA, 2019. Association for Computing Machinery.
\newblock ISBN 9781450368704.
\newblock \doi{10.1145/3341163.3347727}.

\bibitem[Tang et~al.(2021)Tang, Perez-Pozuelo, Spathis, and Mascolo]{simclr}
Chi~Ian Tang, Ignacio Perez-Pozuelo, Dimitris Spathis, and Cecilia Mascolo.
\newblock Exploring contrastive learning in human activity recognition for healthcare, 2021.

\bibitem[Xu et~al.(2021)Xu, Ghosh, Huang, Okhonko, Aghajanyan, Metze, Zettlemoyer, and Feichtenhofer]{xu2021videoclip}
Hu~Xu, Gargi Ghosh, Po-Yao Huang, Dmytro Okhonko, Armen Aghajanyan, Florian Metze, Luke Zettlemoyer, and Christoph Feichtenhofer.
\newblock Videoclip: Contrastive pre-training for zero-shot video-text understanding.
\newblock \emph{arXiv preprint arXiv:2109.14084}, 2021.

\bibitem[Xue et~al.(2022)Xue, Sun, Liu, Fu, Song, Li, and Luo]{xue2022clip}
Hongwei Xue, Yuchong Sun, Bei Liu, Jianlong Fu, Ruihua Song, Houqiang Li, and Jiebo Luo.
\newblock Clip-vip: Adapting pre-trained image-text model to video-language representation alignment.
\newblock \emph{arXiv preprint arXiv:2209.06430}, 2022.

\bibitem[Singer et~al.(2022)Singer, Polyak, Hayes, Yin, An, Zhang, Hu, Yang, Ashual, Gafni, et~al.]{singer2022make}
Uriel Singer, Adam Polyak, Thomas Hayes, Xi~Yin, Jie An, Songyang Zhang, Qiyuan Hu, Harry Yang, Oron Ashual, Oran Gafni, et~al.
\newblock Make-a-video: Text-to-video generation without text-video data.
\newblock \emph{arXiv preprint arXiv:2209.14792}, 2022.

\bibitem[Tevet et~al.(2022)Tevet, Gordon, Hertz, Bermano, and Cohen-Or]{tevet2022motionclip}
Guy Tevet, Brian Gordon, Amir Hertz, Amit~H Bermano, and Daniel Cohen-Or.
\newblock Motionclip: Exposing human motion generation to clip space.
\newblock In \emph{Computer Vision--ECCV 2022: 17th European Conference, Tel Aviv, Israel, October 23--27, 2022, Proceedings, Part XXII}, pages 358--374. Springer, 2022.

\bibitem[Moon et~al.(2022)Moon, Madotto, Lin, Dirafzoon, Saraf, Bearman, and Damavandi]{moon2022imu2clip}
Seungwhan Moon, Andrea Madotto, Zhaojiang Lin, Alireza Dirafzoon, Aparajita Saraf, Amy Bearman, and Babak Damavandi.
\newblock Imu2clip: Multimodal contrastive learning for imu motion sensors from egocentric videos and text.
\newblock \emph{arXiv preprint arXiv:2210.14395}, 2022.

\bibitem[Wang et~al.(2021)Wang, Xing, and Liu]{wang2021actionclip}
Mengmeng Wang, Jiazheng Xing, and Yong Liu.
\newblock Actionclip: A new paradigm for video action recognition.
\newblock \emph{arXiv preprint arXiv:2109.08472}, 2021.

\bibitem[Sun et~al.(2022)Sun, Ke, Rahmani, Bennamoun, Wang, and Liu]{sun2022human}
Zehua Sun, Qiuhong Ke, Hossein Rahmani, Mohammed Bennamoun, Gang Wang, and Jun Liu.
\newblock Human action recognition from various data modalities: A review.
\newblock \emph{IEEE transactions on pattern analysis and machine intelligence}, 45\penalty0 (3):\penalty0 3200--3225, 2022.

\bibitem[Morshed et~al.(2023)Morshed, Sultana, Alam, and Lee]{morshed2023human}
Md~Golam Morshed, Tangina Sultana, Aftab Alam, and Young-Koo Lee.
\newblock Human action recognition: A taxonomy-based survey, updates, and opportunities.
\newblock \emph{Sensors}, 23\penalty0 (4):\penalty0 2182, 2023.

\bibitem[Girdhar et~al.(2023)Girdhar, El-Nouby, Liu, Singh, Alwala, Joulin, and Misra]{girdhar2023imagebind}
Rohit Girdhar, Alaaeldin El-Nouby, Zhuang Liu, Mannat Singh, Kalyan~Vasudev Alwala, Armand Joulin, and Ishan Misra.
\newblock Imagebind: One embedding space to bind them all.
\newblock In \emph{Proceedings of the IEEE/CVF Conference on Computer Vision and Pattern Recognition}, pages 15180--15190, 2023.

\bibitem[Fortes~Rey et~al.(2024)Fortes~Rey, Ray, Xia, Wu, and Lukowicz]{fortes2024enhancing}
Vitor Fortes~Rey, Lala Shakti~Swarup Ray, Qingxin Xia, Kaishun Wu, and Paul Lukowicz.
\newblock Enhancing inertial hand based har through joint representation of language, pose and synthetic imus.
\newblock In \emph{Proceedings of the 2024 ACM International Symposium on Wearable Computers}, pages 25--31, 2024.

\bibitem[Xie et~al.(2018)Xie, Sun, Huang, Tu, and Murphy]{xie2018rethinking}
Saining Xie, Chen Sun, Jonathan Huang, Zhuowen Tu, and Kevin Murphy.
\newblock Rethinking spatiotemporal feature learning: Speed-accuracy trade-offs in video classification.
\newblock In \emph{Proceedings of the European conference on computer vision (ECCV)}, pages 305--321, 2018.

\bibitem[Guo et~al.(2022)Guo, Zou, Zuo, Wang, Ji, Li, and Cheng]{guo2022generating}
Chuan Guo, Shihao Zou, Xinxin Zuo, Sen Wang, Wei Ji, Xingyu Li, and Li~Cheng.
\newblock Generating diverse and natural 3d human motions from text.
\newblock In \emph{Proceedings of the IEEE/CVF Conference on Computer Vision and Pattern Recognition}, pages 5152--5161, 2022.

\end{thebibliography}
\end{document}